# A Novel Deep Learning Method for Textual Sentiment Analysis

Hossein Sadr[*]

Department of Computer Engineering, Rasht Branch, Islamic Azad University, Rasht, Iran

Sadr@qiau.ac.ir

Mozhdeh Nazari Solimandarabi

Department of Computer Engineering, Rasht Branch, Islamic Azad University, Rasht, Iran

Mojdeh.Nazari@qiau.ac.ir

Mir Mohsen Pedram

Department of Electrical and Computer Engineering, Faculty of Engineering, Kharazmi University, Tehran, Iran

Pedram@khu.ac.ir

Mohammad Teshnehlab

Industrial Control Center of Excellence, Faculty of Electrical and Computer Engineering, K. N. Toosi University Tehran, Iran

Teshnehlab@etd.kntu.ac.ir

*Abstract*—Sentiment analysis is known as one of the most crucial tasks in the field of natural language processing and Convolutional Neural Network (CNN) is one of those prominent models that is commonly used for this aim. Although convolutional neural networks have obtained remarkable results in recent years, they are still confronted with some limitations. Firstly, they consider that all words in a sentence have equal contributions in the sentence meaning representation and are not able to extract informative words. Secondly, they require a large number of training data to obtain considerable results while they have many parameters that must be accurately adjusted. To this end, a convolutional neural network integrated with a hierarchical attention layer is proposed which is able to extract informative words and assign them higher weight. Moreover, the effect of transfer learning that transfers knowledge learned in the source domain to the target domain with the aim of improving the performance is also explored. Based on the empirical results, the proposed model not only has higher classification accuracy and can extract informative words but also applying incremental transfer learning can significantly enhance the classification performance.

## 1- Introduction

During the past few years, a large amount of texts containing people's opinions, sentiments, attitudes, and emotions have been rapidly produced due to the explosive growth of social media. Considering the fact that collecting and analyzing such a large amount of unstructured data is not possible, it has been tried to provide an efficient method to collect and process them automatically. That automatic process of text analysis and computational linguistics with the aim of extracting subjective information existing in the text is known as sentiment analysis[1, 2]. Sentiment analysis is considered as one of the most active research areas in the field of natural language processing which tries to classify a piece of text containing opinions based on its polarity and determine whether an expressed opinion about a specific topic, event or product is positive or negative[3, 4].

In recent years, deep neural networks, due to their remarkable results, have attracted many researchers in this filed. However, despite the fact that deep learning models have been quite effective in the field of sentiment analysis, they still suffer from over-abstraction problems [5, 6]. It means that these models can only clarify the polarity of the document and are not able to provide a depth understanding of the text such as identifying the main word that resulted in the polarity classification [7, 8]. It must be also taken into consideration that all parts of a text are not equally important and some words in a sentence have more impact in specifying the whole meaning of the text. Moreover, deep learning models, unlike human brains, cannot pay more attention to the salient part of a text which results in a reduction in their effectiveness [9].

Recently, a new direction of deep learning models has emerged that tries to simulate the attention mechanism found in human brains. Attention mechanism tries to focus on the more important part of a text (as the human brain while reading) and neglects the less important parts[2, 10]. On the other hand, convolutional neural networks have been widely used for the aim of sentiment analysis and have achieved significant progress due to their ability in extracting robust and abstract features from the input [11-13]. In this regard, we decided to integrate the attention mechanism and convolutional neural networks to present a powerful model for sentiment analysis of texts. The intuition behind our model is that all words in a sentence are not equally important and our model can identify the most informative words and phrases of sentences using attention layer by considering the context when phrase-word level sentiment labels are not available[9, 14, 15].

Furthermore, convolutional neural networks require a large number of training data to accurately train the model as well as they need many parameters to be tuned preciously. To this end, transfer learning, as a sub-domain of machine learning, has obtained considerable attention in recent years. The focus of transfer learning is on learning knowledge from the source domain and then apply it to the target domain [16, 17]. In fact, the impact of transfer learning is visible when the training set is not large enough to train the model efficiently. Therefore, the model can be trained in large size datasets and then transformed to the target domain which can lead to performance enhancement. In this regard, we also decided to investigate the effect of transfer learning in our proposed model and explore the sensitivity of the parameters in the source domain. The main contribution of this paper is as follows:

- It is the first time that convolutional neural network and hierarchical attention layer are integrated to mimic the human brain aiming to focus on salient words and phrases in the text for the task of sentiment analysis. The proposed model is able to provide insight into which words carry more valuable information and contribute to the classification decision considering the context.
- The effect of transfer learning with and without training the model on the target domain is explored. Based on the obtained result, applying transfer learning has remarkably enhanced the overall performance.
- An extensive set of experiments were conducted in this paper to demonstrate the sensitivity of parameters and obtain the optimal values for training the model.

The rest of the paper is organized as follows. Related studies are briefly described in section 2. Details of the proposed model are completely explained in section3. Datasets, model configuration, training, and experimental results are described in section 4 in detail. Conclusions and directions for future research are indicated in Section 5.

## 2- RELATED WORK

In recent years, deep learning has been the center of attention and it can be said that it has made a revolution in various filed especially natural language processing. Nowadays, the effect of deep learning in various tasks like text classification, document summarization, machine translation, language modeling, and etc. is completely obvious and sentiment analysis is one of the important aspects of natural language processing that obtained considerable improvement using deep neural networks.

Deep learning contains various networks such as Convolutional Neural Network (CNN) [18], Recursive and Recurrent Neural Network (RNN) [19, 20], and Deep Belief Network (DBN) [21] that have been extensively used in the field of sentiment analysis. In this regard, Kim et al. [18] conducted a series of experiments based on one layer convolutional neural network for this aim. They trained their models on pre-trained vectors derived from Word2Vec embedding model. They also employed multi-channel representation and various filer sizes and achieved comparable results. Against modeling sentences at the word level, Zhang et al. [11] presented a character level CNN for text classification that showed significant enhancement in classification accuracy. Moreover, Kalchbrenner et al. [22] a dynamic CNN that utilized dynamic k-max pooling. While their model was able to handle input sentences of variable lengths, it could efficiently capture short and long-term dependencies. Yin and Schutze presented a multichannel variable size CNN that employed combinations of various word embedding techniques as input [23].

In the following, Tai et al. [24]employed Long Short Term Memory (LSTM) network integrated with some complex units for sentiment analysis. They also conducted more experiments on 2 layers and bidirectional LSTM and achieved significant results. Following a similar line of research, Kuta et al. [25] proposed tree structure gated recurrent neural network which was inspired by tree structure LSTM and adaptation of Gated Recurrent Unit (GRU) to recursive model.

Besides these networks, a semi-supervised model known as the Recursive neural network has been also employed for the task of sentiment analysis which uses continuous word vectors as input and hierarchical structure. In this regard, Socher et al. [19] introduced a model, known as MV-RNN, that employed both matrix and vector with the aim of representing words and phrases in the tree structure. Recursive Neural Tensor Network (RNTN) is another network in this field that was proposed by Socher et al. [26] where the tensor-based compositional matrix was used instead of matrix representation for all nodes in the tree structure.

In spite of the fact that deep neural networks have achieved significant results in the field of sentiment analysis, they are still confronting with some limitations. One of their general pitfalls is that they consider all words in the sentences equally and are not able to focus on salient parts of the text [7]. To fill this lacuna, the attention mechanism has been recently adopted in many tasks of natural language processing especially sentiment analysis due to its strength in providing an effective interpretation of the text. In this regard, Yang et al. [7] modified the RNN by adding weight that played attention role for the aim of text classification. Wang et al.[27] also proposed an attention-based LSTM network that could focus on various parts of the sentences. It must be taken into consideration that despite promising results of applying attention mechanism on deep neural networks, only a few studies have been conducted in the field of sentiment analysis.

On the other hand, another challenge that deep neural networks are commonly confronted with refers to the lack of training data. In fact, deep neural networks require a large number of training data to be able to accurately train the model and as the number of data is increased, their performance is also enhanced [28]. The lack of available labeled training data has yielded to the emergence of a new concept known as transfer learning. Transfer learning is used when the training set is not large enough to efficiently train the model. Therefore, the model is trained on a large dataset known as the source domain and then is transferred to the target domain which can significantly enhance the performance. Although transfer learning has been extensively used in the field of image processing, its application in natural language processing, especially sentiment analysis, is still limited.

In this regard, Krizhevsky and Lee [29], presented the efficiency of transferring low-level neural layers in different tasks. In a similar study [30], the impact of transferring high-level layers in a deep neural network from source dataset to smaller size target dataset was investigated. However, it is worth mentioning that the effect of transfer learning for the task of sentiment analysis has been rarely explored [9, 31].

Considering the mentioned challenges, we decided to propose a novel CNN integrated with attention layer for the aim of sentiment analysis in this paper. Despite previous studies, the proposed model applies attention mechanism after convolutional layer to extract informative words existing in the sentences by assigning a higher weight to them which leads to the creation of the new representation of word vectors. Moreover, in order to improve the overall performance of the proposed model, we decided to employ transfer learning. In fact, the proposed model is first trained on a large dataset and then is transferred to the target dataset.

### 3- METHODOLOGY

The proposed model of this paper consists of two main processes. The first process is a slight variant of the classical Convolutional Neural Network (CNN) presented by Kim et al.[18] which employs a hierarchical attention mechanism on convolutional neural network to emphasize words that have a significant effect in determining the meaning of sentences. The focus of the second process is on transfer learning which tries to store knowledge learned from a source domain and apply it to various but related target domains.

**3-1- Process 1:** Convolutional neural network integrated with a hierarchical attention mechanism

The learning process flow of convolutional neural network integrated with a hierarchical attention mechanism is presented in Fig.1 which contains four layers. Firstly, by performing word embedding, word vectors of input sentences are extracted and are then joined to form the initial input matrix for CNN. Secondly, the CNN model is trained. As the training is completed, feature maps extracted from similar filter sizes are merged and fed to the attention layer as a new matrix in the third layer. In the following, by extracting the informative words by assigning a higher weight to them using attention mechanism and aggregating their representation to the previous features extracted by convolution neural network, new sentence vector are formed. Finally, new word vectors are entered into a fully connected network and classification is performed. More detailed mathematical deduction about each layer is provided in the following.

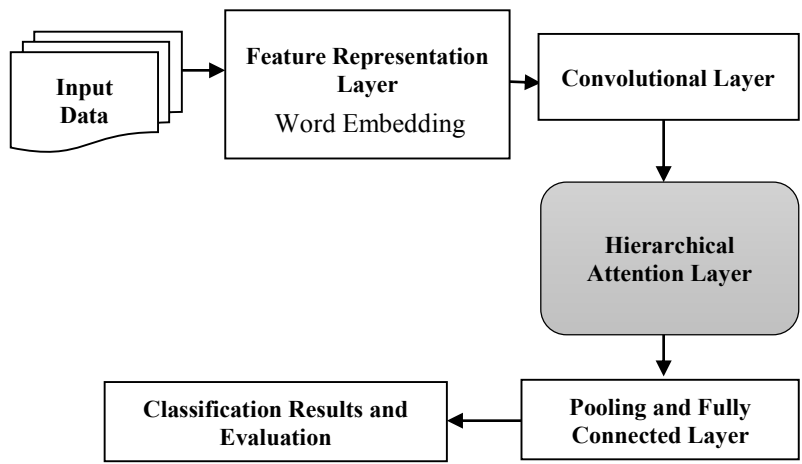

Fig.1. Learning Process in convolutional neural network integrated with a hierarchical attention mechanism.

*3-1-1- Feature representation layer*

The convolutional neural network requires a sentence matrix as an input where each row represents a word vector. If the dimensionality of word vector is $d$ and the length of a given sentence is $s$, the dimensionality of sentence matrix would be $s \times d$ where padding is set to zero before the first word and after the last word in the sentence. Setting the padding to zero makes the number of times that each word is included in receptive filed during the convolution the same without considering the word position in the sentence. As a result, the sentence matrix is denoted by $A \in \mathcal{R}^{s \times d}$. In this paper, various word embedding techniques including *Word2Vec* [32], *Glove* [33] and *FastText* [34] are employed to form the input matrix.

*3-1-2- Convolution layer*

To produce new features, the convolutional operation must be applied to the sentence matrix. According to the fact that the sequential structure of a sentence has an important effect in determining its meaning, it is sensible to choose filter width equal to the dimensionality of word vectors ($d$). In this regard, only the height of filters ($h$), known as region size, can be varied.

Considering $A \in \mathcal{R}^{s \times d}$ as a sentence matrix, convolution filter $H \in \mathcal{R}^{h \times d}$ is applied on $A$ to produce its submatrix as new feature $A[i:j]$. As the convolution operation is applied repeatedly on the matrix of $A$, $O \in \mathcal{R}^{s-h+1 \times d}$ as the output sequence is achieved (Eq. 1).

$$O_i = w \; o \; A[i:i+h-1] \quad (1)$$

Here $i = 1, \dots, s - h + 1$ and $o$ is the dot product between the convolution filter and submatrix. Bias term $b \in \mathcal{R}$ and an activation function are also added to each $O_i$. Finally, feature maps $C \in \mathcal{R}^{s-h+1}$ are generated (Eq.2).

$$C_i = f(O_i + b) \quad (2)$$

*3-1-3- Attention Layer*

Whereas it is believed that all words in a sentence do not contribute equally to represent the meaning of a sentence, there is a need for a mechanism to emphasize such words that have more impact on the meaning of the sentences. For this aim, we decided to apply an attention mechanism on feature maps extracted from the previous layer. In this regard, feature maps that are extracted from the same filter size are aggregated and form a new matrix.

Suppose that in the convolution layer, $M$ different region sizes are considered and for each of them $m$ different filters are employed. Therefore, after applying $H_{ij} \in \mathcal{R}^{h_i \times d}$ filters on sentence matrix $A$ where $i = 1,2,\dots,M$ and $j = 1,2,\dots,m$, $M \times m$ feature map is obtained. By concating feature maps extracted from the same filter size, a new sentence matrix $X_i \in \mathcal{R}^{n \times m}$ (Eq.3) is obtained. Where $n$ is the number of words and each element of this matrix represents the feature extracted from the input using filters with the same size.

$$X_i = \begin{bmatrix} \bar{x}_{1,1} & \cdots & \bar{x}_{1,m} \\ \vdots & \ddots & \vdots \\ \bar{x}_{n-c_i+1,1} & \cdots & \bar{x}_{n-c_i+1,m} \end{bmatrix} \quad (3)$$

The objective of the attention mechanism is to assign specific weight to each row for extracting informative parts of the sentence. For this aim, firstly, new word matrix $X_i$ is fed through a single layer perceptron using $w \in \mathcal{R}^{m \times d}$ and $U_i \in \mathcal{R}^{n-h_i+1 \times d}$ is obtained as a hidden representation of $X_i$ (Eq.4).

$$U_i = \tanh(\bar{X}_i W + b) \quad (4)$$

In the following, the importance of each word is measured as the similarity of $U_i$ with a content vector $u \in \mathcal{R}^{d \times 1}$ to achieve the normalized importance weight $a_i \in \mathcal{R}^{n-h_i+1 \times 1}$ using *Softmax* (Eq.5).

$$a_i = \text{softmax}(U_i u) \quad (5)$$

Content vector $u$ tries to specify informative words. Notably, $u$ is set to zero in the beginning to compute the same weight for various rows in the matrix of $X_i$ and it is learned during the training process. After that, $\bar{X}_i$ (a new representation of $X_i$) is computed by multiplying each element of $a_i$ to its corresponding row in $X_i$ matrix (Eq.6).

$$\bar{X}_i = a_i \; o \; X_i \quad (6)$$

Generally, $\bar{X}_i$ is a new representation of $X_i$ while the attention mechanism is applied to it in order to specify the informative words.

The whole process of attention layer is schematically presented in Fig. 2. As it can be clearly seen, after merging feature maps extracted from the same filter sizes, $X_i$ matrix is created. Then, by applying a single layer perceptron, a new representation of $X_i$ known as $U_i$ is created. In the following, the normalized importance weight $a_i$, indicating the importance of each word, is computed as the similarity between $U_i$ and content vector $u$ which is a hyper-parameter and is tuned during the training process. Finally, $\bar{X}_i$ is a new representation of $X_i$ is achieved by multiplying each element of $a_i$ to its corresponding row in $X_i$. Generally, applying the attention mechanism leads to extracting informative words and assigning more weight to them. The overall process of attention layer is depicted in Fig.2.

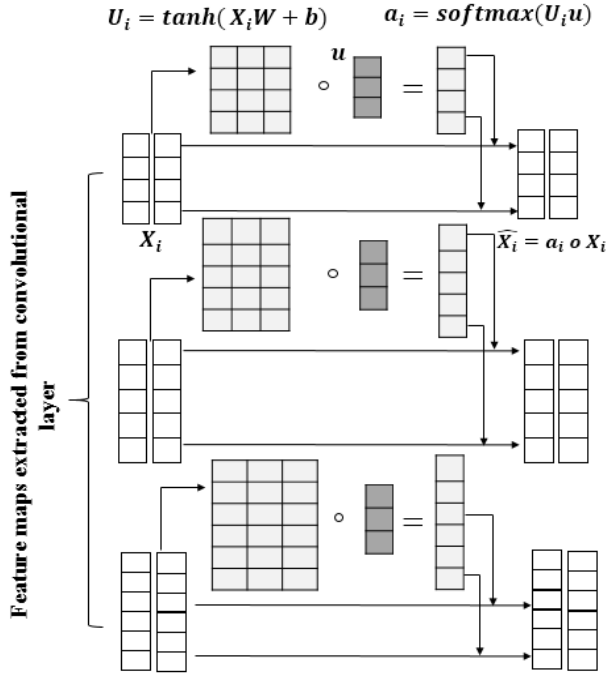
Fig.2. The overall process of attention layer

*3-1-4- Pooling and fully connected layer*

While various feature maps according to different filter sizes are generated, a pooling function is required to induce fixed size vectors. Various strategies such as average pooling, minimum pooling, and maximum pooling can be used for this aim and the idea behind them is to capture the most important feature from each feature map and reduce dimensionality. Generated features from pooling layer from each filter are concated into a feature vector $o_i$. The feature vector is passed to a fully connected *Softmax* layer to specify the final classification. In other words, *Softmax* determines the probability distribution over all sentiment categories and is calculated as follows (Eq.7).

$$P_i = \frac{\exp(o_i)}{\sum_{j=1}^{c} \exp(o_j)} \quad (7)$$

To clarify the difference between the real sentiment distribution $\widehat{P}_i(C)$ and the distribution achieved from the model $P_i(C)$, cross-entropy as the loss function is employed (Eq.8).

$$Loss = -\sum_{s \in T}\sum_{i=1}^{V} \widehat{P}_i(C) \log(P_i(C)) \quad (8)$$

Where $T$ is the training set and $V$ is the sentiment categories. Stochastic Gradient Descent (SGD) is also used for end to end training of the model.

**3-2- Process 2:** Transfer learning

According to the fact that increasing the number of training data has a significant effect on the performance of deep neural networks[35], we decided to employ transfer learning with the aim of combining two different domains to improve classification accuracy. The transfer learning process that is used in this paper for training the proposed model is depicted in Fig. 3 and Fig.4. As it is clear, firstly, the convolutional neural network integrated with a hierarchical attention mechanism is trained on the source domain based on the process flow that was shown in Fig. 1 and then the trained model is transferred to the target domain. Transferring the trained model also includes two different processes. In the first one, convolutional neural network integrated with a hierarchical attention mechanism does not learn from the target domain. It means that the trained model is only tested on the target domain. While in the second process, the model is trained on the target domain to incrementally learn and update its knowledge. It means that in the second process the model is trained on both source and target domain and can use their combination to increase the classification performance. Finally, this new trained model evaluates the data in the target domain.

**4- EXPERIMENTS**

**4-1- Dataset**

In order to have a comprehensive investigation of the effectiveness of the proposed model, standard datasets for the aim of sentiment analysis were used in our experiments. While the focus of this paper is on exploring the impact of transfer learning, different datasets are used as source and target domains in our experiments. In other words, transfer learning is used when the target domain is not large enough to be able to train the model accurately. In this regard, we decided to use a large dataset as a source domain and then train the proposed model on a smaller dataset known as the target domain. A brief description of the source and target domains that are used in this paper are provided in the following.

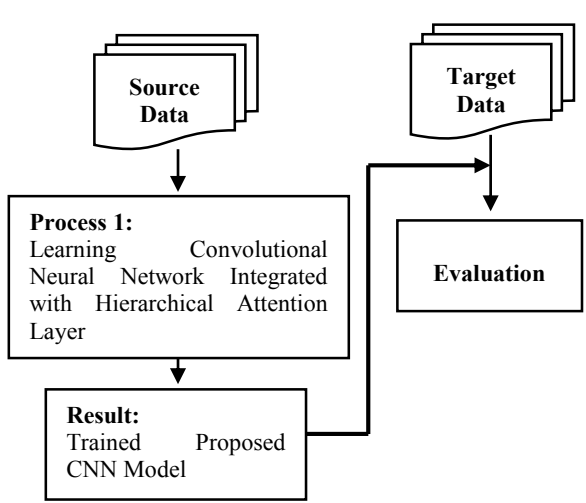 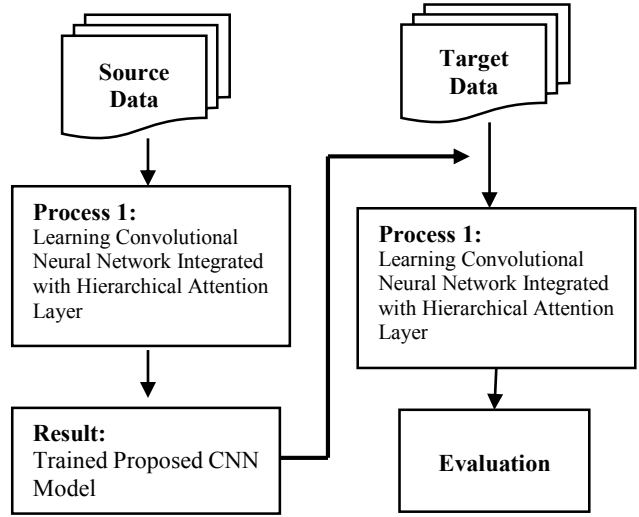

Fig.3. Transfer learning without training the model on the target domain.

Fig.4. Transfer learning with training the model on the target domain

- **Source domain:** Amazon Review
  This dataset contains reviews about Amazon products that were collected by Zhang et al. [11]. This dataset has 2 classes (*AMZ-2*) and 5 classes (*AMZ-5*) versions[1].
- **Target domain:** Stanford Sentiment Treebank
  This is the dataset that is commonly used for sentiment classification which also contains 2 classes (*SST-2*) and 5 classes (*SST-5*) versions. In fact, this it is the extended version of the MR dataset [36] which also contains train/dev/test sets and fine-grained labels[2].

It is worth mentioning that the reason behind using the Amazon review dataset as a source domain refers to its large size that makes it a suitable source and another reason refers to the low degree of semantic similarity between these two datasets. According to the experiments conducted with Zhang et al. [28], as two datasets are lesser semantically similar to each other, applying transfer learning demonstrates greater performance. The reason behind using Stanford Sentiment Treebank as a target domain refers to the fact that this dataset is used in the majority of research conducted for the task of sentiment analysis which can provide this opportunity to compare the proposed model with a wide range of existing models. The summary statistics of these datasets is presented in Table 1.

Table 1. Summary Statistics of used datasets

|  | **Dataset** | **C** | **L** | **S** | **V** |
|---|---|---|---|---|---|
| $D_S$ | AMZ-5 | 5 | 84 | 3650000 | 1057296 |
|  | AMZ-2 | 2 | 82 | 3000000 | 1112820 |
| $D_T$ | SST-5 | 5 | 18 | 11855 | 17836 |
|  | SST-2 | 2 | 19 | 9613 | 16185 |

*$D_S$: Source domain, $D_T$: Target Domain, *C*: Number of classes, *L*: Average sentence length, *S*: Number of sentences, *V*: Vocabulary size

**4-2- Model Configuration**

Generally, one of the downsides of the convolutional neural networks refers to their free number of hyper-parameters which require practitioners to determine the exact model architecture. While the hyper-parameters' values have a significant impact on the performance of deep neural networks, we decided to optimize the proposed model hyper-parameters on the source domain and then apply the optimal parameters on the target domain. In this regard, we first used the CNN configuration proposed by Zhang et al. [37] and tried to carry out extensive sets of experiments to obtain the optimal values for the proposed model. The used baseline configuration is presented in

---

[1] https://goo.gl/bm0IkT
[2] http://nlp.stanford.edu/sentiment/

Table 2. Notably, 10-fold cross-validation where 10% of training data was randomly selected as a test set was performed and each experiment was repeated 5 times and the average results are reported.

Table 2. Baseline configuration

| Hyper-parameters | Value |
|---|---|
| Word embedding | Random |
| Filter region size | 3,4,5 |
| Number of filters | 512 |
| Dropout rate | 0.5 |
| Activation Function | ReLU |

*4-2-1- Word embedding effect*

One of the interesting properties of sentence classification models refers to their ability to use distributed representation of words as input while they can also use various pre-trained word vectors to form initial input. In this section, the sensitivity of the proposed model with respect to different input representations is explored. In this regard, we used 5 input variations in our experiments.

Firstly, we used random initialized word vectors as input (*Random*). Then, the model was trained using pre-trained word vectors that were obtained using Word2Vec algorithm (*Word2Vec-Static*). The third one refers to using pre-trained word vectors that were obtained using Word2Vec algorithm and were also updated during the training process (*Word2Vec-nonStatic*). Next, the model was trained using word vectors obtained from a combination of Word2Vec and random initialized vectors (*2-channel*). Finally, we used a combination of pre-trained word vectors achieved from Word2Vec, Glove, and FastText as an input while they were also updated during the training of the model (*4-channel*).

Noteworthy, along the word vectors training process, the dimension of word vectors was considered as 200 and window size was set to 3. Word vectors were updated based on learning rate of {0.1, 0.05, 0.01}. It must be noted that skip-gram structure was utilized in Word2Vec [32] and FastText [34]models while Glove[33] used unigram structure.

Table 3. The effect of the word embedding on the performance of the proposed model

| Word Embedding | Accuracy % | |
|---|---|---|
| | AMZ-2 | AMZ-5 |
| Random | 92.44 | 57.31 |
| Word2Vec-Static | 92.85 | 57.9 |
| Word2Vec-nonStatic | 93.12 | 58.7 |
| 2-channel | 93.6 | 58.6 |
| 4-channel | **93.88** | **58.91** |

Based on the obtained results (Table 3), *Random* word embedding has the lowest classification accuracy among all variations. Better performance of other variations can be also attributed to the employment of pre-trained vectors that can solve the semantic sparsity problem to some degree. In other words, it can be claimed that pre-trained vector representation has a great effect on the performance of the proposed model. Moreover, considering the fact that *Word2Vec-Static* embedding has the lowest classification accuracy besides *Random* embedding, it can be stated that updating word vectors during the training process can yield to obtain higher performance without considering if the word vectors were previously trained or not. Finally, *4-channel* embedding has the highest classification accuracy. In this regard, we used *4-channel* word embedding to train the proposed model on the target domain.

*4-2-2- Filter region size effect*

In order to investigate the effect of filter region size, various numbers of region sizes were explored while the other parameters were kept constant. According to previous studies that demonstrated the priority of multiple region sizes in comparison to the single region size, we only used multiple region sizes in our experiments. As can be seen in Table 4, various regions size has a great impact on the performance of the model and the best obtained result is different with baseline value which indicates that the greatest accuracy is obtained while the multiple region size was set as (4, 5, 6). In this regard region size (4,5,6) was used in the target domain.

### 4-2-3- Number of filter effect

Again, in this set of experiments, other configurations were held constant and we only changed the number of filters in each region. Based on the obtained results (Table 5), it is clear the number of filters has also considerable impact on the performance of the proposed model and the proposed model obtained the highest accuracy while the number of filters was set to 300. Therefore, we used 300 filters to train the model on the target domain.

Table 4. The effect of the region size on the performance of the proposed model

| Region size | Accuracy % | |
|---|---|---|
| | AMZ-2 | AMZ-5 |
| (3,4,5) | 92.44 | 57.31 |
| **(4,5,6)** | **92.98** | **57.65** |
| (6,7,8) | 91.8 | 56.85 |
| (8,9,10) | 90.78 | 55.85 |
| (9,10,11) | 90.45 | 55.32 |
| (14,15,16) | 90.30 | 55.24 |
| (3,4,5,6) | 91.10 | 56.07 |
| (6,7,8,9) | 90.17 | 55.14 |
| (7,7,7,7) | 92.25 | 57.03 |
| (7,7,7) | 91.43 | 56.13 |

Table 5. The effect of the number of filters on the performance of the proposed model

| Number of filters | Accuracy % | |
|---|---|---|
| | AMZ-2 | AMZ-5 |
| 125 | 91.5 | 56.32 |
| 256 | 91.78 | 56.54 |
| **300** | **92.83** | **57.74** |
| 512 | 92.44 | 57.31 |
| 450 | 90.25 | 55.34 |
| 640 | 90.85 | 55.95 |

### 4-2-4- Regularization effect

Dropout is a technique that is generally used for the aim of regularization and avoiding overfitting. In fact, by using dropout, a number of neurons are randomly selected to be ignored during training which has a great impact on the performance of the model. In this set of experiments, different dropout rates in the range of 0.1 to 0.9 were used to find the optimal rate. Based on the obtained results (Table 6), the highest accuracy was obtained when the dropout rate was around 0.6 and therefore this dropout rate was used to train the model on the target domain.

Table 5. The effect of the regularization on the performance of the proposed model

| Dropout rate | Accuracy % | |
|---|---|---|
| | AMZ-2 | AMZ-5 |
| 0.1 | 91.54 | 56.73 |
| 0.2 | 91.40 | 56.32 |
| 0.3 | 91.93 | 56.57 |
| 0.4 | 91.8 | 56.23 |
| 0.5 | 92.44 | 57.31 |
| **0.6** | **92.89** | **57.68** |
| 0.7 | 91.48 | 56.41 |
| 0.8 | 92.08 | 57.12 |
| 0.9 | 92.28 | 57.43 |

*4-2-5- Activation function effect*

Considering the fact that activation function plays the role of a guide for every input in filters, it is then specified as one of the prominent parts in convolutional neural networks. Different activation functions such as *ReLU, Tanh, SoftPlus* and *linear* can be used convolutional neural networks that each of them has its own characteristics. Based on the obtained results (Table 7), the *ReLU* function outperformed the other activation function and therefore it was used to train the model on the target domain.

Table 7. The effect of the activation function on the performance of the proposed model

| Activation Function | Accuracy % | |
|---|---|---|
| | AMZ-2 | AMZ-5 |
| Tanh | 91.45 | 57.12 |
| Softplus | 80.25 | 40.43 |
| **ReLU** | **92.4** | **57.31** |
| Linear | 90.31 | 56.25 |

### 4-3- Results

While the proposed model of this paper is divided into two sections, the obtained results are also categorized into two parts. In this regard, we first investigate the effect of employing a hierarchical attention mechanism on convolutional neural network and then consider the impact of using transfer learning. More details are provided in the following.

*4-3-1- Classification results*

To create a baseline and provide a fair comparison among the proposed model and another state of the arts, firstly, we only trained our proposed model on the target domain without considering transfer learning. Accuracy comparison of our model against other existing models is provided in Table 8. As it is clear, the proposed model is compared with a wide range of deep neural networks and machine learning models. Based on the obtained results, the proposed model has slightly superior performance compared to existing models and therefore it can be considered as a good option for being used in transfer learning.

Table 8. Results of experiments on SST1 and SST2 dataset

| Model | Accuracy % | |
|---|---|---|
| | SST-2 | SST-5 |
| NB[26] | 81.8 | 41 |
| BiNB[26] | 83.1 | 41.9 |
| SVM [12] | 79.4 | 40.7 |
| WordVec-AVE [12] | 80.1 | 32.7 |
| CNN-1 layer [22] | 77.1 | 37.4 |
| CNN-non static[18] | 87.2 | 48 |
| CNN-multichannel[18] | 88.1 | 47.4 |
| DCNN [22] | 86.8 | 48.5 |
| LSTM[24] | 85.2 | 46.2 |
| Bi-LSTM[24] | 87.5 | 49.1 |
| Tree-LSTM[24] | 88.0 | 51.0 |
| Tree-GRU[38] | 88.6 | 50.5 |
| Tree-GRU+ attention[38] | 89.0 | 51.0 |
| LSTM+RNN attention[27] | 86.1 | 48.0 |
| RecRNN[26] | 82.4 | 43.2 |
| RNTN[26] | 85.4 | 45.7 |
| MVRNN[19] | 82.9 | 44.4 |
| **Proposed model** | **90.57** | **51.31** |

It must be noted that the optimal values, discussed in the previous section, were used to train the model and the result corresponding to other existing models are taken from their original papers. Moreover, the accuracies were obtained using the available test data *ADADELTA* update rule which is an algorithm in the family of stochastic gradient descent and can be trained over shuffled mini-batches was used for optimization. Each experiment was repeated for five times and average results reported.

To have a broader view of the performance of the proposed model, more analysis has been also carried out. In fact, the aim of applying an attention mechanism on CNN is to focus on more relevant words. Without the attention mechanism, CNN might also work well and assign a high and low weight to important and not important words respectively without considering the context. While the importance of a particular word is highly dependent on the context, the goal of the proposed model is to capture context-dependence importance.

To clarify the performance of the proposed model in recognition of the importance of the word based on the context, the distribution of the attention weight of words good and bad from test split of SST1 dataset is presented in Fig.5 (a, b). According to the distribution, the assigned attention weight is in the scale of 0 to 1. This specifies that the potential of the proposed method is in capturing diverse context and assigning context dependent weight.

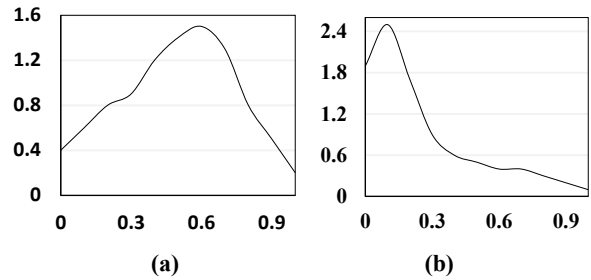

Fig.5. Distribution of attention weight of words *good* (a) and *bad* (b)

In order to have more comprehensive analysis, the distributions of words *good* and *bad* are plot according to the rating of reviews in Fig. 6 and 7 (a)-(e) corresponding to the ratings 1 to 5 respectively. Notably, the Fig. 6 is related to word *good* and the Fig. 7 is related to the word *bad*. As can be clearly seen, in reviews with rating 1, the words *good* and *bad* have the lowest and highest weight respectively. In the following, as the rating is enhanced, the weight distribution for the word *good* is increased while it is decreased for the word *bad*. It indicates that positive words such as *good* have a more important role in higher rating reviews while negative words such as *bad* have more effect in lower rating reviews. In other words, it can be stated that for the word *good* as the rating goes higher, the distribution also shifts higher. In contrast, the word *bad* has a higher distribution in poor ratings while it decreases for good ones. This indicates that the proposed model is able to capture the importance of words without considering the context

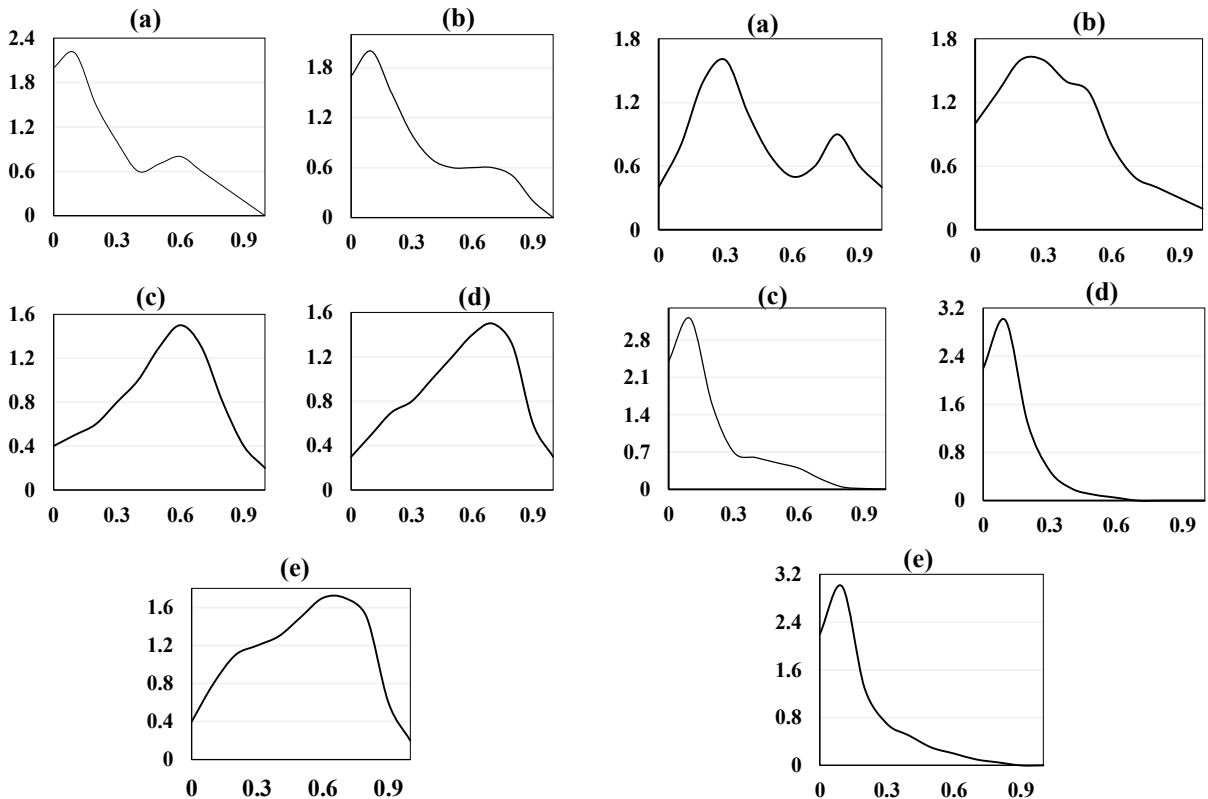

Fig.6. Distribution of attention weight of words *good* according to the ratings (1 to 5)

Fig.7. Distribution of attention weight of words *bad* according to the ratings (1 to 5)

*4-3-2- Transfer learning results*

One of the big challenges that deep neural networks are confronted with refers to the lack of enough labeled training data. In fact, the performance of deep neural networks is highly dependent on the number of data and increasing the number of training data has a significant effect on their performance. Transfer learning here comes to help increase the size of the training set. Following a similar line of research, we first trained our model on the Amazon review dataset and transfer it to Stanford Sentiment Treebank.

As it was previously mentioned, two different learning process is used in the proposed model (Fig.3 and Fig.4). In one of them, the proposed model is directly used for sentiment classification in the target domain and in the other one, the model is incrementally trained on the target domain using the optimal values that previously discussed. In order to make a comparison between these transferring processes, the accuracy of the proposed model without and with incremental learning in the target domain (Source domain: AMZ2, Target domain: SST2) is presented in Table 9. As it is clear, the accuracy of the model is very low when it is directly used for classification which indicates that the knowledge obtained in the source domain is not enough to be applied in the target domain. On the other hand, when the model is incrementally learned in the target domain, the accuracy is significantly improved.

Table 9. Transfer learning performance comparison with and without incremental learning

| Description | Accuracy (%) |
|---|---|
| Transfer learning without incremental learning | 74.3 |
| Transfer learning with incremental learning | 91.25 |

In order to provide more analysis of the impact of transfer learning on sentiment classification, the results of employing transfer learning with incremental learning on all variations of source and target domains are presented in Table 10. As it is clear, employing transfer learning has generally enhanced the overall classification performance. Specifically, although AMZ5→SST2 and AMZ2→SST5 have different number of classes demonstrated the highest performance. In fact, it can be stated that employing transfer learning has remarkably enhanced the performance of the proposed model which can be yielded to the large size of the source domain and richer embedding which helps the model to learn contextual information better.

Table 10. Accuracy (%) of transfer learning on different variations of source and target domains

| $D_T$ | → SST-2 | | →SST-5 | |
|---|---|---|---|---|
| $D_S$ | AMZ-2 | AMZ-5 | AMZ-2 | AMZ-5 |
| Proposed model | 91.25 | **92.38** | **53.63** | 52.16 |

## 5- CONCLUSION

The contribution of this paper is twofold. Firstly, a new convolution neural network integrated with the attention mechanism for the aim of sentiment analysis is proposed that tries to consider the context to determine the polarity of sentences. In fact, it can not only predict the sentimental label of a sentence but also find informative words that effectively contribute to predicting the overall classification decision. Generally, the proposed model progressively forms sentence vectors by aggregating informative words vectors achieved from attention layer into feature maps extracted from the convolutional layer and employ these new generated features for classification. Secondly, the sensitivity of the proposed model parameters is comprehensively investigated and the effect of transfer learning is explored.

According to the empirical results and due to the best of our knowledge, the proposed convolution neural network integrated with the attention mechanism significantly outperformed other existing models. Moreover, employing transfer learning has greatly improved the classification accuracy. In fact, the proposed model obtained an accuracies of 90.57% and 51.31% on SST-2 and SST-5 datasets respectively without transfer learning. On the other hand, by applying transfer learning the classification accuracy has increased and the obtained accuracies were about 92.38% and 53.63% on SST-2 and SST-5 datasets respectively.

Following a similar line of research, the proposed model using transfer learning can be performed in another target domain or for other natural language processing tasks. Applying transfer learning on other deep neural networks is also worth exploring.